\documentclass[runningheads]{llncs}

\usepackage[T1]{fontenc}
\usepackage{bm}
\usepackage{graphicx}
\usepackage{amsmath,amssymb}
\usepackage{booktabs}
\usepackage{multirow}
\usepackage{xcolor}
\usepackage{hyperref}
\usepackage{cleveref}
\usepackage{wrapfig}
\usepackage{mathabx}

\newcommand{\boldparagraph}[1]{\vspace{0.0cm}\noindent{\bf #1.}} 

\graphicspath{{figures/}}

\usepackage{tikz}
\newcommand\copyrighttext{%
  \scriptsize This preprint has not undergone peer review or any post-submission improvements or corrections. The Version of Record of this contribution is published in \textit{Proceedings of the 27th International Conference on Artificial Intelligence in Education (AIED), Seoul, South Korea, 27 Jun -- 3 July 2026}, and is available online at [DOI will be added later].
  \smallskip
  
  \textcopyright\ 2025. Please cite this article as follows: G. Li, L. Chen, C. Tang, B. Ma, Y. Jiang, D. Deguchi, T. Yamashita, A. Shimada: \textit{Capture-Calibrate-Coach: A Graph-Based Framework for Knowledge Monitoring Estimation and Adaptive Feedback}. International Conference on Artificial Intelligence in Education (AIED), Springer Nature, 2026. (Book: Artificial Intelligence in Education.) DOI: DOI: \href{DOI will be added later}{DOI will be added later}.}
\newcommand\copyrightnotice{%
\begin{tikzpicture}[remember picture,overlay]
\node[anchor=north,yshift=-24pt] at (current page.north) {\fbox{\parbox{\dimexpr\textwidth-\fboxsep-\fboxrule\relax}{\copyrighttext}}};
\end{tikzpicture}%
}

\begin{document}

\title{Capture-Calibrate-Coach: A Graph-Based Framework for Knowledge Monitoring Estimation and Adaptive Feedback}
\titlerunning{3C: Graph-Based Knowledge Monitoring Estimation}
\authorrunning{G.\,Li et al.}
\author{
  Gen Li\inst{1} \and
  Li Chen\inst{2} \and
  Cheng Tang\inst{1} \and
  Boxuan Ma\inst{1} \and
  Yuncheng Jiang\inst{3} \and \\
  Daisuke Deguchi\inst{4} \and
  Takayoshi Yamashita\inst{5} \and
  Atsushi Shimada\inst{1}
}

\institute{
    Kyushu University, Fukuoka, Japan \\
\email{\{gen.li, atsushi\}@limu.ait.kyushu-u.ac.jp} \and
    Osaka Kyoiku University, Osaka, Japan \\
\email{chen-l68@cc.osaka-kyoiku.ac.jp} \and
    South China Normal University, Guangzhou, China \and
    Nagoya University, Nagoya, Japan \and
    Chubu University, Kasugai, Japan \\
}

\maketitle

\begin{abstract}
Effective learning support requires understanding not only what learners know but also how accurately they perceive their own understanding. This metacognitive dimension, known as knowledge monitoring, fundamentally influences self-regulated learning, yet this dimension remains underexplored in current systems. This paper introduces the \textit{Capture-Calibrate-Coach} (3C) framework for adaptive learning support. The \textit{Capture} phase extracts learners' perceived knowledge states from open-ended self-reports to construct a heterogeneous graph linking learners and knowledge concepts. The \textit{Calibrate} phase applies a heterogeneous graph neural network to infer latent perceived states for concepts not explicitly mentioned, enabling systematic knowledge monitoring assessment. The \textit{Coach} phase classifies learners into five metacognitive patterns and delivers personalized feedback addressing both knowledge gaps and calibration errors. Evaluation with 684 students demonstrates 85.21\% AUC in predicting latent perceived states, significantly outperforming baseline methods. A user study with 47 participants shows positive reception of feedback quality, with participants particularly valuing concrete feedback on knowledge gaps and actionable study guidance. These findings advance AI-based learning support toward metacognitive teammates that foster accurate self-awareness while supporting knowledge growth.
\copyrightnotice
\end{abstract}


\section{Introduction}
\vspace{-4pt}

Accurate self-assessment of one's own understanding is a cornerstone of successful learning. When learners can reliably distinguish between what they know and what they do not know, they can allocate study time more strategically, seek help when needed, and avoid the pitfall of overconfidence~\cite{zimmerman2002becoming}. This metacognitive capacity, known as \textit{knowledge monitoring}, has been consistently linked to improved academic outcomes across diverse educational contexts~\cite{lingel2019metacognition,was2014discrimination}. 

With the proliferation of Learning Management Systems (LMS) and Intelligent Tutoring Systems (ITS), data-driven approaches to learning support have become increasingly sophisticated~\cite{mousavinasab2021intelligent}. These systems primarily evaluate learners' knowledge states through test performance, enabling personalized recommendations for content review and practice. However, evaluating knowledge states alone provides an incomplete picture of learner needs. Two problematic phenomena frequently observed in educational practice highlight this limitation: \textit{overconfidence}, where learners believe they understand material they actually do not, and \textit{underconfidence}, where learners doubt their understanding despite possessing adequate knowledge~\cite{bol2012calibration}.
These calibration errors stem from deficiencies in knowledge monitoring. Therefore, learning support systems should consider not only learners' knowledge states but also knowledge monitoring.

Traditional approaches to evaluating knowledge monitoring rely on structured confidence judgments, where learners rate their confidence for each assessment item~\cite{prokop2020calibration}. However, this approach requires pre-defining assessment items, limiting flexible application across diverse curricula. In contrast, open-ended learning reflections, a core practice in the self-reflection phase of Self-Regulated Learning (SRL), are implemented in many educational institutions \cite{goker2016use,li2025single}. Through such reflection practices, learners produce self-reports that freely articulate their understanding, reflecting more authentic metacognitive processes compared to passive responses to predefined options. However, a fundamental challenge arises: learners do not mention every assessed knowledge concept in their open-ended self-reports derived from reflections, leaving their perceived states for many concepts unknown. We term these unobserved perceptions Latent Perceived States (LPS), and their inference is essential for comprehensive knowledge monitoring assessment (KMA).

To address these challenges, we propose a \textit{Capture-Calibrate-Coach} (3C) framework that integrates test logs and learning reflections to simultaneously evaluate learners' knowledge states and knowledge monitoring ability. The framework operates through the following three phases.
First, in the \textit{Capture} phase, we extract learners' explicitly stated perceptions from self-reports and construct a heterogeneous graph linking learners to knowledge concepts. Large language models (LLMs) facilitate this extraction process, enabling scalable analysis of open-ended text.
Second, in the \textit{Calibrate} phase, we formulate the inference of LPS as a link prediction problem on the heterogeneous graph. A Heterogeneous Graph Neural Network (HGNN) leverages both learner-concept interactions and semantic relationships among concepts to infer whether learners perceive themselves as understanding concepts they did not explicitly mention, then evaluates knowledge 
monitoring ability using Signal Detection Theory (SDT) metrics.
Third, in the \textit{Coach} phase, we classify learners into five metacognitive patterns based on the alignment between their perceived states and actual test performance. Following Hattie and Timperley's feedback model~\cite{hattie2007power}, we deliver personalized feedback addressing both knowledge gaps and calibration errors. 

This paper addresses the following research questions:
\begin{itemize}
\vspace{-4pt}
    \item \textbf{RQ1:} To what extent can LLMs accurately extract knowledge concepts and identify learner perceptions from educational materials and self-reports?
    \item \textbf{RQ2:} How effectively can HGNN-based link prediction infer learners' latent perceived states for knowledge concepts not explicitly mentioned?
    \item \textbf{RQ3:} How do learners and educators perceive the usefulness of adaptive feedback based on knowledge monitoring patterns?
\vspace{-4pt}
\end{itemize}

\section{Related Work}
\vspace{-4pt}

\boldparagraph{Knowledge Monitoring Assessment} Knowledge monitoring refers to the metacognitive ability to accurately judge one's own knowledge states, essential for effective self-regulated learning~\cite{tobias1996assessing,zimmerman2002becoming}. Signal Detection Theory provides a principled framework for quantifying this ability by cross-tabulating perceived states against actual performance, yielding metrics such as discriminability ($d'$), sensitivity, and specificity~\cite{maniscalco2012signal,barrett2013measures}. Was et al.~\cite{was2014discrimination} connected this SDT framework with knowledge monitoring assessment, showing that $d'$ captures discrimination ability while avoiding confounds in earlier metrics like gamma. Subsequent work emphasized that single metrics often conflate discrimination with response bias, advocating for multi-metric approaches~\cite{schraw2009conceptual}. However, existing approaches rely on structured confidence judgments for pre-defined assessment items. Open-ended learning reflections allow learners to freely articulate their understanding~\cite{goker2016use}, yet methods to extract knowledge monitoring from such unstructured text remain underdeveloped.

\boldparagraph{Graph-Enhanced Modeling in Education}
Graph-based representations effectively encode relationships in educational contexts. Educational knowledge graphs structure relationships among 
learning content, including prerequisite dependencies~\cite{liu2016learning,chen2018knowedu}. Recent approaches leverage LLMs to automate knowledge graph construction from educational materials~\cite{meyer2023llm,li2024llm}, and student dialogue data in collaborative settings~\cite{chen2024three,11194887}. Learner-content interaction graphs further enable personalized recommendations~\cite{Frej2024}. Graph Neural Networks (GNNs) have advanced knowledge tracing by modeling dependencies between exercises and knowledge concepts. For example, hierarchical contrastive graph frameworks capture both local and global structural information~\cite{huang2025hcgkt}, and GKT-CD infers mastery levels through learner-concept interactions~\cite{zhang2021gkt}. Recent work applies graph-based methods to learner reflections by transforming them into Personal Knowledge Graphs and extracting structural patterns correlated with learning outcomes~\cite{li2025reflections}. However, these approaches focus on inferring actual knowledge states. The inference of perceived states remains unexplored.

\section{Proposed Method}
\vspace{-4pt}

\begin{figure*}[t]
\centering
\includegraphics[width=\textwidth]{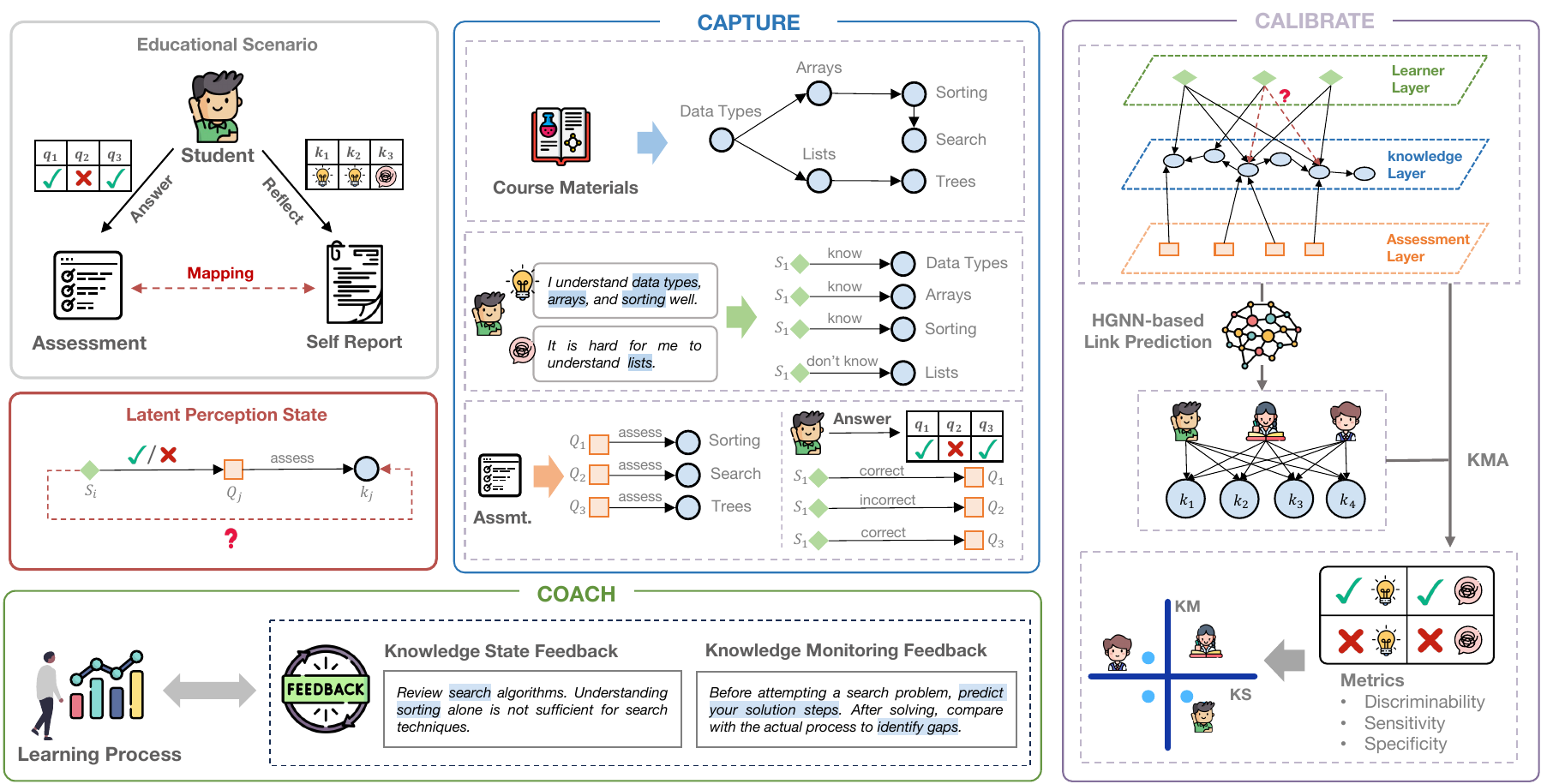}
\caption{\textbf{Overview of the 3C framework.}
The \textit{Capture} phase extracts knowledge concepts and prerequisite relationships from course materials, maps learner perceptions from  self-reports, and records assessment responses. The \textit{Calibrate} phase constructs a heterogeneous graph with learner, knowledge, and assessment layers, then applies HGNN-based link prediction to infer latent perceived states (marked with ``?'') and assesses knowledge monitoring. The \textit{Coach} phase computes knowledge monitoring metrics and delivers personalized feedback addressing both knowledge gaps and calibration errors.}
\label{fig:framework}
\vspace{-12pt}
\end{figure*}

This section presents the Capture-Calibrate-Coach (3C) framework for estimating learners' knowledge monitoring by integrating test logs and self-reports. Figure~\ref{fig:framework} provides an overview of the framework.

\subsection{Problem Formulation}
Consider an educational setting where learners $\mathcal{S} = \{s_1, \ldots, s_n\}$ participate in a course covering knowledge concepts $\mathcal{K} = \{k_1, \ldots, k_m\}$. Each learner $s_i$ engages in two activities:
   
\noindent \textbf{Learning reflections}: Learners $s_i$ produce open-ended self-reports articulating their understanding, mentioning knowledge concepts they perceive they \textit{know} ($\mathcal{K}_i^+$) and \textit{don't know} ($\mathcal{K}_i^-$).
   
\noindent \textbf{Assessments}: Learners respond to assessment items $\mathcal{Q} = \{q_1, \ldots, q_t\}$, where each $q_j$ primarily assesses concept $k_j \in \mathcal{K}$. The outcomes are recorded as $\mathcal{R}_i = (r_{i,1}, \ldots, r_{i,t})$ with $r_{i,j} \in \{0, 1\}$ indicating incorrect or correct.

A critical challenge is that learners do not necessarily mention every assessed knowledge concept in their self-reports. Let $\mathcal{K}_Q = \{k_j \mid q_j \in \mathcal{Q}\}$ denote assessed concepts. For many learners, $\mathcal{K}_Q \not\subseteq (\mathcal{K}_i^+ \cup \mathcal{K}_i^-)$, leaving perceived states for some concepts unobserved (i.e., Latent Perceived States (LPS)). However, not mentioning a concept does not necessarily mean \textit{don't know}, as learners may have reported understanding of prerequisite or neighboring concepts. Therefore, given learners' open-ended self-reports, assessment outcomes, and learning context (e.g., course materials), our objectives are: (1) Infer LPS to complete learners' perceived state profiles, enabling accurate estimation of knowledge monitoring ability. (2) Provide adaptive feedback based on learners' knowledge states and knowledge monitoring ability.

\subsection{Capture: Heterogeneous Graph Construction}
The Capture phase extracts structured information from course materials, assessment data, and open-ended self-reports to construct a heterogeneous graph that encodes knowledge structure, assessment-concept associations, and learners' perceived states.

\boldparagraph{Concept and Relation Extraction} 
We extract knowledge concepts $\mathcal{K} = \{k_1, \ldots, k_m\}$ and their 
relationships from course materials, focusing on prerequisite relationships 
that are fundamental for learners to navigate learning paths~\cite{aytekin2024ace}. The extraction produces a directed acyclic graph representing the hierarchical knowledge structure, with edges $\mathcal{E}_{K\text{-}K} \subseteq \mathcal{K} \times \mathcal{K}$ where $(k_i, k_j) \in \mathcal{E}_{K\text{-}K}$ indicates that $k_i$ is a prerequisite for $k_j$.

\boldparagraph{Assessment-Concept Linking}
To establish the association between assessment items and knowledge concepts, 
we leverage LLMs to analyze each item's text against the extracted concepts, 
linking each assessment item $q_j \in \mathcal{Q}$ to its primarily assessed 
concept $M(q_j) \in \mathcal{K}$.
This forms edges $\mathcal{E}_{Q\text{-}K} = \{(q_j, M(q_j)) \mid q_j \in \mathcal{Q}\}$. This step may be bypassed when items are pre-labeled with target concepts during assessment design.

\boldparagraph{Learner Perception Extraction}
From each learner's self-report, we extract concepts they perceive they \textit{know} and those they perceive they \textit{don't know} based on contextual cues. The extracted concepts form two sets for each learner $s_i$: 
$\mathcal{K}_i^+$ containing concepts they perceive they \textit{know}, and 
$\mathcal{K}_i^-$ containing concepts they perceive they \textit{don't know}.
These establish learner-concept edges $\mathcal{E}_{S\text{-}K} = \{(s_i, k) \mid k \in \mathcal{K}_i^+\}$ and $\mathcal{E}_{S\text{-}K}^- = \{(s_i, k) \mid k \in \mathcal{K}_i^-\}$.

All extraction and linking processes above can be facilitated using LLMs, 
and the resulting outputs can be reviewed by domain experts. As a result, the Capture phase produces a heterogeneous graph $G_C = (V, E)$ comprising three node types (i.e., learners $V_S$, knowledge concepts $V_K$, and assessment items $V_Q$) connected by concept prerequisites ($\mathcal{E}_{K\text{-}K}$), assessment-concept links ($\mathcal{E}_{Q\text{-}K}$), and learner-concept perception edges ($\mathcal{E}_{S\text{-}K}$, $\mathcal{E}_{S\text{-}K}^-$). However, learners do not mention every assessed concept in their self-reports. For concepts in $\mathcal{K}_Q$ not covered by $\mathcal{K}_i^+ \cup \mathcal{K}_i^-$, the perceived states remain unknown. These \textit{latent perceived states} are the target of inference in the Calibrate phase.

\subsection{Calibrate: LPS Inference and Knowledge Monitoring Evaluation}
The Calibrate phase infers LPS for concepts not explicitly mentioned in learners' self-reports and evaluates knowledge monitoring by comparing perceived states with actual test performance.

\boldparagraph{Perception Subgraph Construction}
We formulate LPS inference as a link prediction problem on a perception subgraph. From the heterogeneous graph $G_C$, we extract a subgraph $G_P = (V_S \cup V_K, E_P)$ containing learner nodes $V_S$ and knowledge concept nodes $V_K$. The edge set $E_P$ comprises:
\begin{itemize}
\vspace{-5pt}
    \item Learner-concept edges ($\mathcal{E}_{S\text{-}K}$): connecting 
    learners to concepts they perceive they know
    \item Concept-concept edges ($\mathcal{E}_{K\text{-}K}$): prerequisite 
    relationships between concepts
\vspace{-5pt}
\end{itemize}
Importantly, edges representing concepts perceived as \textit{don't know}
($\mathcal{E}_{S\text{-}K}^-$) are not included in $G_P$.
In this formulation, edge existence indicates \textit{know} perception, 
while edge absence indicates \textit{don't know} perception.
Thus, inferring whether a learner perceives an unmentioned concept as \textit{know} or \textit{don't know} becomes predicting whether an edge should 
exist between them.

\boldparagraph{Explicit-Informed Negative Sampling}
Since edges for \textit{don't know} perceptions are excluded from $G_P$, 
these explicit reports ($\mathcal{K}_i^-$) provide reliable negative 
supervision for training the link prediction model.
We design an Explicit-Informed Negative Sampling (EINS) strategy that 
prioritizes these explicit negatives over randomly sampled ones. For each learner $s_i$, the negative sample set is constructed as:
\begin{equation}
    \mathcal{N}_i = \text{Sample}(\mathcal{K}_i^-, N_e) \cup 
    \text{Sample}\left(\mathcal{K}_Q \setminus (\mathcal{K}_i^+ \cup \mathcal{K}_i^-), 
    \lfloor \rho \cdot N_e \rfloor \right)
\end{equation}
where $N_e$ negative samples are drawn from explicit negatives, 
and additional implicit negatives are sampled from unmentioned concepts.
The ratio $\rho$ balances explicit and implicit supervision: 
smaller $\rho$ emphasizes reliable explicit negatives, 
while larger $\rho$ encourages generalization to unmentioned concepts.
By prioritizing explicit \textit{don't know} reports, the model learns more 
accurate boundaries of learners' self-perception.
The model is trained by minimizing binary cross-entropy loss over 
positive edges in $G_P$ and negative samples in $\mathcal{N}_i$.

\boldparagraph{Link Prediction with HGNN}
We employ Heterogeneous Graph Neural Networks (HGNNs) for link prediction, 
as the perception subgraph contains different node types and edge types 
that require type-specific processing.
HGNNs handle this heterogeneity through type-specific message passing:
\begin{equation}
    h_v^{(l+1)} = \text{Aggregate}_{\tau(v)}^{(l)}\left(h_v^{(l)}, 
    \{h_u^{(l)}, r : u \in N_r(v), \forall r \in R\}\right)
\end{equation}
where $\tau(v)$ denotes node $v$'s type, $N_r(v)$ represents neighbors 
connected by relation type $r$, and $R$ is the set of all relation types. After computing node embeddings $h_s$ for learners and $h_k$ for knowledge 
concepts, the probability that learner $s$ perceives concept $k$ as ``know'' 
is predicted as:
\begin{equation}
    \hat{y}_{s,k} = \sigma(f(h_s, h_k))
\end{equation}
where $\sigma$ is the sigmoid function and $f$ computes the compatibility 
between learner and concept embeddings. Using the trained model, we infer perceived states for unmentioned concepts. For each learner $s_i$ and concept $k_j \in \mathcal{K}_Q \setminus 
(\mathcal{K}_i^+ \cup \mathcal{K}_i^-)$:
\begin{equation}
    k_j \in \begin{cases} 
        \hat{\mathcal{K}}_i^+ & \text{if } \hat{y}_{s_i, k_j} \geq \theta \\
        \hat{\mathcal{K}}_i^- & \text{if } \hat{y}_{s_i, k_j} < \theta 
    \end{cases}
\end{equation}
where $\theta$ is a threshold parameter. Each learner's perceived states are then completed by combining explicit reports with inferred states: $\mathcal{K}_i^+ \cup \hat{\mathcal{K}}_i^+$  for \textit{know} and $\mathcal{K}_i^- \cup \hat{\mathcal{K}}_i^-$ for \textit{don't know}.

\boldparagraph{Knowledge Monitoring Assessment} With inferred LPS, we construct complete perception profiles for each learner, covering all assessed concepts with their perceived states (\textit{know} or \textit{don't know}). These profiles enable systematic knowledge monitoring assessment. Following prior work on knowledge monitoring assessment, we apply Signal Detection Theory (SDT) metrics to quantify knowledge monitoring ability~\cite{tobias2009importance,smith2019knowledge,was2014discrimination}.

\begin{wraptable}{r}{0.53\textwidth}
\vspace{-30pt}
\centering
\caption{\textbf{Contingency table for KMA}}
\label{tab:sdt}
\small
\begin{tabular}{l|cc}
\toprule
 & Correct & Incorrect \\
\midrule
Know & Hit (A) & False Alarm (B) \\
Don't Know & Miss (C) & Correct Rejection (D) \\
\bottomrule
\end{tabular}
\vspace{-10pt}
\end{wraptable}

For each assessment item, we cross-tabulate perceived state against 
actual performance, yielding four SDT categories (Table \ref{tab:sdt}).
We derive three metrics that jointly characterize knowledge monitoring:

\begin{itemize}
\vspace{-5pt}
    \item \textbf{Discriminability} ($d' = z(A/(A{+}C)) - z(B/(B{+}D))$): 
    measures overall ability to distinguish known from unknown concepts, 
    where $z(\cdot)$ is the inverse normal CDF. 
    Higher $d'$ indicates better discrimination 
    ability~\cite{barrett2013measures,was2014discrimination}.
    \item \textbf{Sensitivity} ($A/(A{+}C)$): 
    measures the ability to correctly identify known concepts among 
    all correctly answered items.
    Low sensitivity indicates \textit{underconfidence}, where 
    learners doubt knowledge they actually possess~\cite{lingel2019metacognition,was2014discrimination}.
    \item \textbf{Specificity} ($D/(B{+}D)$): 
    measures the ability to correctly identify unknown concepts among 
    all incorrectly answered items.
    Low specificity indicates \textit{overconfidence}, where 
    learners believe they understand material they actually 
    do not~\cite{lingel2019metacognition,was2014discrimination}.
\vspace{-5pt}
\end{itemize}

\noindent While $d'$ quantifies overall discrimination ability, sensitivity 
and specificity reveal the \textit{direction} of miscalibration. 
This characterization enables the Coach phase to classify learners 
into metacognitive patterns and deliver differentiated feedback.

\subsection{Coach: Learner Classification and Adaptive Feedback}

The Coach phase classifies learners into metacognitive patterns based on the alignment between their perceived states and actual test performance, then delivers personalized feedback.

\boldparagraph{Learner Classification} Based on test performance and knowledge monitoring metrics , we classify learners into five patterns (Table~\ref{tab:patterns}):

\begin{itemize}
\vspace{-4pt}
    \item \textbf{Well Calibrated}(\textit{WC}): Learners with strong performance and accurate self-assessment. They correctly identify what they \textit{know} and \textit{don't know}.
    \item \textbf{Aware of Limitations}(\textit{AL}): Learners with weak performance but accurate self-assessment. Despite knowledge gaps, they recognize what they do not know.
    \item \textbf{Underconfident}(\textit{UC}): Learners with strong performance but low sensitivity. They doubt knowledge they actually possess, potentially leading to unnecessary review of mastered content.
    \item \textbf{Overconfident}(\textit{OC}): Learners with weak performance and low specificity. They believe they understand material they actually do not, risking premature termination of study.
    \item \textbf{Liberal Criterion}(\textit{LC}): Learners with strong performance but a tendency to respond \textit{know} across items. In SDT terms, they adopt a liberal response criterion, shifting the decision threshold to more \textit{know} responses~\cite{kantner2012response}. This results in high sensitivity but low specificity despite adequate knowledge.
\vspace{-4pt}
\end{itemize}

\begin{table}[t!]
\centering
\caption{\textbf{Learner patterns based on performance and knowledge monitoring.} Learners are first split by performance level, then by 
discriminability ($d'$). For low-$d'$ learners, sensitivity and specificity further distinguish the direction of miscalibration.}
\vspace{-8pt}
\label{tab:patterns}
\small
\setlength{\tabcolsep}{8pt}
\resizebox{0.95\textwidth}{!}{
\begin{tabular}{lccc}
\toprule
\midrule
\textbf{Pattern} & \textbf{Performance} & \textbf{$d'$} & \textbf{Additional Criterion} \\
\midrule
Well Calibrated (\textit{WC}) & High & High & — \\
Aware of Limitations (\textit{AL}) & Low & High & — \\
\midrule
Underconfident (\textit{UC}) & High & Low & Low Sensitivity \\
Overconfident (\textit{OC}) & Low & Low & Low Specificity \\
Liberal Criterion (\textit{LC}) & High & Low & High Sensitivity \\
\midrule
\bottomrule
\end{tabular}}
\vspace{-10pt}
\end{table}

\boldparagraph{Adaptive Feedback Generation} To structure feedback according to each learner pattern, we adopt Hattie and Timperley's~\cite{hattie2007power} feedback framework, which organizes effective feedback around three questions:

\begin{itemize}
\vspace{-4pt}

\item \textbf{Feed Up} (\textit{Where am I going?}) presents the learner's current position within the five metacognitive patterns, the connection between current position and target goal of achieving \textit{Well Calibrated}, and the priority between addressing knowledge gaps and improving knowledge monitoring.

\item \textbf{Feed Back} (\textit{How am I going?}) presents correct and incorrect knowledge concepts, related past errors identified via the knowledge graph, and the four SDT categories for each assessed concept.

\item \textbf{Feed Forward} (\textit{Where to next?}) employs an LLM to generate actionable advice. To address knowledge gaps, it provides explanations for incorrect concepts, consolidation-oriented guidance for low-performing learners to review and strengthen foundational understanding, and reinforcement-oriented guidance for high-performing learners to challenge and extend mastered knowledge. To improve knowledge monitoring, it provides pattern-specific strategies designed based on empirical findings from knowledge monitoring theories and classic learning theories. The key theoretical foundations are summarized in Table~\ref{tab:feedback_theory}.
\vspace{-4pt}
\end{itemize}

\begin{table}[h!]
\vspace{-8pt}
\centering
\caption{\textbf{Learning theories informing Feed Forward design.} Each theory provides strategies targeting specific learner patterns, guiding the LLM to generate pattern-specific actionable advice in the Feed Forward component.}
\vspace{-8pt}
\label{tab:feedback_theory}
\resizebox{\columnwidth}{!}{
\begin{tabular}{l|l|l|l}
\toprule
\midrule
\textbf{Learning Theory} & \textbf{Target} & \textbf{Core Strategy} & \textbf{Ref.} \\
\midrule
\textit{Self-regulated learning} & \textit{All} & Planning, monitoring, reflective regulation & \cite{zimmerman2002becoming} \\
\textit{Anxiety-cognitive capacity} & \textit{All} & Low-stakes assessment, self-pacing & \cite{tobias1985test} \\
\textit{Knowledge monitoring} & \textit{UC}, \textit{OC}, \textit{LC} & Strategic help seeking, fading scaffolding & \cite{tobias2009importance} \\
\textit{Self-verification} & \textit{OC} & Self-explanation, expectation-outcome comparison & \cite{butler1995feedback} \\
\textit{Cognitive load theory} & \textit{UC}, \textit{OC}, \textit{LC} & Attention guidance, prioritized feedback cues & \cite{sweller1988cognitive} \\
\textit{Depth of processing} & \textit{WC}, \textit{AL} & Apply to new problems, why-and-how question & \cite{craik1972levels} \\
\midrule
\bottomrule
\end{tabular}}
\vspace{-1cm}
\end{table}

\section{Evaluation}
\vspace{-4pt}
We evaluate the 3C framework through three studies addressing each research question:
(1) accuracy of LLM-based extraction,
(2) effectiveness of HGNN-based LPS inference, and
(3) user perception of adaptive feedback.
\subsection{Experimental Setup}
\vspace{-4pt}
\boldparagraph{Dataset} We conducted experiments using data from a \textit{Foundations of Computing} course at a university. The dataset comprises 684 students across five classes, with the course spanning 14 weeks. The curriculum covered 23 topics including information representation, logic circuits, cryptography, algorithms, data structures, and machine learning fundamentals. As described in the problem formulation, learners engaged in two activities after each lecture: (1) open-ended self-reports articulating their understanding, and (2) assessment tests evaluating their knowledge states.

\boldparagraph{Ethical Considerations} All learner data were anonymized to protect privacy, and our experiments complied with institutional ethical guidelines. Learners provided informed consent for their data to be used in this research.
\subsection{LLM-Based Heterogeneous Graph Construction (RQ1)}
We evaluate the accuracy of LLM-based extraction for three tasks in the Capture phase: (1) extracting knowledge concepts and prerequisite relationships from course materials, (2) linking assessment items to knowledge concepts, and (3) extracting learner perceptions from self-reports.

\boldparagraph{Experimental Protocol} We employed \textit{GPT-5.2} for all extraction tasks and report means over 3 runs to ensure stable estimates. In addition, for knowledge graph construction, we adopted the Graphusion strategy~\cite{10.1145/3701716.3717821}, which performs topic-wise extraction followed by fusion across topics. Three domain experts independently annotated the ground truth for each task. Particularly, for perception extraction, due to the large volume of learner data, we randomly sampled 432 self-reports for manual annotation. Inter-annotator disagreements were resolved through discussion to produce the final ground truth.

\begin{table}[b!]
\vspace{-8pt}
\caption{\textbf{LLM-based extraction accuracy in 
the Capture phase.} Few-shot prompting benefits relation extraction, while zero-shot suffices for the remaining tasks.}
\vspace{-8pt}
\centering
\setlength{\tabcolsep}{5pt}
\resizebox{\textwidth}{!}
{
\begin{tabular}{ll|cc|cc}
\toprule
\midrule
\multirow{2}{*}{\textbf{Source}} & \multirow{2}{*}{\textbf{Task}} & \multicolumn{2}{c|}{\textbf{Zero-shot}} & \multicolumn{2}{c}{\textbf{Few-shot}} \\
\cmidrule{3-6}
& & Acc. $\uparrow$ & F1 $\uparrow$ & Acc. $\uparrow$ & F1 $\uparrow$ \\
\midrule
\multirow{2}{*}{Course Materials} 
& Concept Extraction & -- & 75.0 & -- & \textbf{75.5} \\
& Relation Extraction & -- & 70.2 & -- & \textbf{76.2} \\
\midrule
Assessment Items & Assessment Linking & \textbf{89.2} & -- & 87.6 & -- \\
\midrule
\multirow{2}{*}{Self-reports} 
& Perception Extraction (\textit{know}) & -- & \textbf{96.7} & -- & 95.0 \\
& Perception Extraction (\textit{don't know}) & -- & \textbf{94.1} & -- & 91.6 \\
\midrule 
\bottomrule 
\end{tabular}}
\label{tab:extraction_accuracy}
\vspace{-0.6cm}
\end{table}

\boldparagraph{Evaluation Metrics} We adopt standard metrics appropriate for each task. For concept and relation extraction, we report F1 scores measuring the overlap between LLM-extracted and expert-annotated outputs, as these tasks involve set matching. For assessment linking, we report accuracy since each assessment item maps to exactly one primary concept. For perception extraction, we report F1 scores for both \textit{know} and \textit{don't know} categories, as this task involves identifying concept mentions within free-form text. We compare zero-shot and few-shot prompting strategies to examine the effect of demonstration examples.

\boldparagraph{Results}
Table~\ref{tab:extraction_accuracy} summarizes extraction accuracy across all tasks. For concept and relation extraction, few-shot prompting outperformed zero-shot, with improvements of 0.5 and 6.0 percentage points respectively. The gain in relation extraction suggests that prerequisite relationships benefit more from demonstration examples. For assessment-concept linking, zero-shot prompting achieved 89.2\% accuracy, while few-shot prompting showed no improvement (87.6\%). Similarly, for perception extraction, zero-shot prompting achieved 96.7\% F1 for \textit{know} and 94.1\% F1 for \textit{don't know}, outperforming few-shot in both cases. These results indicate that both tasks are relatively straightforward for LLMs when concept definitions are provided. Overall, these results demonstrate that LLMs can reliably automate the heterogeneous graph construction process. For further improvement, post hoc review and refine by instructors may be more practical than crafting demonstrations upfront, particularly for the latter two tasks

Table \ref{tab:dataset} presents the statistics of the heterogeneous graphs constructed through the Capture phase for each class. Each graph comprises three node types (students, concepts, and assessments) connected by four edge types: concept prerequisites ($\mathcal{E}_{K\text{-}K}$), assessment-concept links ($\mathcal{E}_{Q\text{-}K}$), student responses ($\mathcal{E}_{S\text{-}Q}$), and student perceptions ($\mathcal{E}_{S\text{-}K}$).
Notably, each class contains a substantial number of LPS instances (ranging from 1,727 to 3,807), representing learner-concept pairs where perceived states were not explicitly mentioned in self-reports. These LPS instances constitute the inference targets for the Calibrate phase.

\begin{table}[h!] 
\vspace{-8pt}
\centering 
\setlength{\tabcolsep}{.5em}
\caption{\textbf{Statistics of the constructed heterogeneous graphs.} \textcolor{red}{\#LPS} denotes the number of Latent Perceived States requiring inference.} 
\vspace{-8pt}
\label{tab:dataset} 
\begin{tabular}{c|c|c} 
\toprule 
\midrule 
\textbf{Class} & \textbf{Node} & \textbf{Edge} \\ 
\midrule 
FC$_A$ & \begin{tabular}[c]{@{}c@{}} \# student: 109 \\ \# knowledge concept: 211 \\ \# assessment items: 48 \end{tabular} & 
\begin{tabular}[c]{@{}c@{}} \# $\mathcal{E}_{K-K}$: 226 | \# $\mathcal{E}_{Q-K}$: 48 | \# $\mathcal{E}_{S-Q}$: 5,232 \\ \# $\mathcal{E}_{S-K}$: 6,349 | \textcolor{red}{\#LPS: 2,187}\end{tabular} \\ 
\midrule 
FC$_B$ & \begin{tabular}[c]{@{}c@{}} \# student: 181 \\ \# knowledge concept: 211 \\ \# assessment items: 38 \end{tabular} & 
\begin{tabular}[c]{@{}c@{}} \# $\mathcal{E}_{K-K}$: 226 | \# $\mathcal{E}_{Q-K}$: 38 | \# $\mathcal{E}_{S-Q}$: 6,878 \\ \# $\mathcal{E}_{S-K}$: 10,128 | \textcolor{red}{\#LPS: 3,002}\end{tabular} \\ 
\midrule 
FC$_C$ & \begin{tabular}[c]{@{}c@{}} \# student: 101 \\ \# knowledge concept: 211 \\ \# assessment items: 43 \end{tabular} & 
\begin{tabular}[c]{@{}c@{}} \# $\mathcal{E}_{K-K}$: 226 | \# $\mathcal{E}_{Q-K}$: 43 | \# $\mathcal{E}_{S-Q}$: 4,432 \\ \# $\mathcal{E}_{S-K}$: 6,908 | \textcolor{red}{\#LPS: 1,727}\end{tabular} \\ 
\midrule 
FC$_D$ & \begin{tabular}[c]{@{}c@{}} \# student: 148 \\ \# knowledge concept: 211 \\ \# assessment items: 42 \end{tabular} & 
\begin{tabular}[c]{@{}c@{}} \# $\mathcal{E}_{K-K}$: 226 | \# $\mathcal{E}_{Q-K}$: 42 | \# $\mathcal{E}_{S-Q}$: 6,258 \\ \# $\mathcal{E}_{S-K}$: 9,396 | \textcolor{red}{\#LPS: 2,634}\end{tabular} \\ 
\midrule 
FC$_E$ & \begin{tabular}[c]{@{}c@{}} \# student: 145 \\ \# knowledge concept: 211 \\ \# assessment items: 50 \end{tabular} & 
\begin{tabular}[c]{@{}c@{}} \# $\mathcal{E}_{K-K}$: 226 | \# $\mathcal{E}_{Q-K}$: 50 | \# $\mathcal{E}_{S-Q}$: 7,300 \\ \# $\mathcal{E}_{S-K}$: 6,918 | \textcolor{red}{\#LPS: 3,807}\end{tabular} \\ 
\midrule 
\bottomrule 
\end{tabular}
\vspace{-0.8cm}
\end{table}

\subsection{LPS Inference Accuracy (RQ2)}

We evaluate the effectiveness of HGNN-based link prediction for inferring Latent Perceived States (LPS) in the Calibrate phase.

\boldparagraph{Baselines and Experimental Setup}
We compared our approach against baseline methods, covering both homogeneous and heterogeneous graph as input:
\begin{itemize}
\vspace{-5pt}
    \item \textbf{Random Guesser (RG)}: A simple baseline that randomly assigns probabilities to potential student-knowledge links.
    
    \item \textbf{GAT} \cite{velikovi2017graph}: Graph Attention Network, a homogeneous graph neural network that incorporates attention mechanisms between nodes and their neighbors.
    
    \item \textbf{GCN} \cite{kipf2016semi}: Graph Convolutional Network, a homogeneous graph neural network that performs neighborhood aggregation through spectral convolutions.
    
    \item \textbf{LP} \cite{huang2020combining}: Label Propagation, a semi-supervised learning algorithm that propagates labels through graph edges.
\vspace{-5pt}
\end{itemize}
Our method employs HAN~\cite{wang2019heterogeneous} based HGNN with the Explicit-Informed Negative Sampling (EINS) strategy described in Section~3. We formulate LPS inference as a link prediction task: predicting whether an edge should exist between a learner and a concept in the perception subgraph $G_P$. Therefore, we use commonly adopted ROC-AUC as the evaluation metric, which measures the model's ability to discriminate between \textit{know} and \textit{don't know} perceptions. For each class, we split the data into 80\% for training and 20\% for testing, and report means over 30 random trials to ensure stable estimates.

\boldparagraph{Results} Table~\ref{tab:lps_accuracy} presents the LPS inference accuracy across all five classes. Our method demonstrates robust performance across all classes (i.e., from 82.33\% to 87.23\%), achieving the highest AUC (i.e., Ave. 85.21\%).

\begin{table}[h]
\vspace{-8pt}
\caption{\textbf{Performance (AUC\%) comparison of methods across five classes.} Columns $\mathcal{E}_{S-K}$ and $\mathcal{E}_{K-K}$ indicate which edge types are used as input.}
\vspace{-8pt}
\centering
\setlength{\tabcolsep}{0.62em}
\begin{tabular}{l|cc|ccccc|c}
\toprule
\midrule
\textbf{Methods} & \multicolumn{2}{c|}{\textbf{Edges}} & \multicolumn{5}{c|}{\textbf{AUC} $\uparrow$} & \\
& $\mathcal{E}_{S-K}$ & $\mathcal{E}_{K-K}$ &\textit{FC$_A$} & \textit{FC$_B$} & \textit{FC$_C$} & \textit{FC$_D$} & \textit{FC$_E$}& \textit{Ave.}\\
\midrule
RG & \checkmark & - & 50.00 & 50.00 & 50.00 & 50.00 & 50.00 & 50.00 \\
GCN & \checkmark & - & 77.91 & 76.97 & 79.18 & 79.53 & 76.45 & 78.01 \\
GAT & \checkmark & - & 75.44 & 74.71 & 75.81 & 77.18 & 73.89 & 75.41 \\
LP & \checkmark & \checkmark & 81.30 & 81.85 & 80.68 & 82.85 & 81.47 & 81.63 \\
\midrule
 Ours & \checkmark & \checkmark & 86.52 & 85.17 & 84.81 & 87.23 & 82.33 & 85.21 \\
 \hspace{.5em} $\drsh$ w/o EINS & \checkmark & \checkmark &$\drsh$ 76.52& 76.49& 78.47&79.49 & 77.58& 77.71 \\
\midrule
\bottomrule
\end{tabular}
\label{tab:lps_accuracy}
\vspace{-10pt}
\end{table}

\boldparagraph{Effect of heterogeneous graph structure}
Comparing methods that use only learner-concept edges ($\mathcal{E}_{S-K}$) versus those that additionally leverage concept-concept edges ($\mathcal{E}_{K-K}$), we observe consistent improvements.
LP (81.63\%) outperformed both GCN (78.01\%) and GAT (75.41\%) by incorporating prerequisite relationships between concepts.
Our method further improved upon LP by 3.58 percentage points, demonstrating the advantage of learning expressive node representations through heterogeneous graph neural networks.

\boldparagraph{Effect of EINS}
The ablation study (Ours w/o EINS) reveals that the Explicit-Informed Negative Sampling strategy contributes substantially to performance.
Without EINS, our method achieves only 77.71\% AUC, which is even lower than LP.
This 7.50 percentage point improvement confirms the importance of utilizing learners' explicit \textit{don't know} reports as more reliable negative supervision.

\subsection{User Perception of Adaptive Feedback (RQ3)}

We evaluate how learners and educators perceive the usefulness of adaptive feedback generated by the Coach phase.

\boldparagraph{Feedback Generation}
Following the adaptive feedback framework described in Section~3, we generate personalized feedback according to Hattie and Timperley's model~\cite{hattie2007power} based on each learner's metacognitive pattern. Learners are classified into five patterns using median splits on test performance and knowledge monitoring metrics ($d'$, sensitivity, specificity). We employed \textit{GPT-5.2} to generate the \textit{Feed Forward} component.

\boldparagraph{Procedure}
We recruited 47 participants (36 university students and 11 instructors) who provided informed consent prior to the study.
Participants first received instruction on the key concepts underlying the feedback, including knowledge monitoring, the SDT-based four-category classification (Hit, False Alarm, Miss, Correct Rejection), and the five learner patterns.
They then reviewed the generated feedback content for learners across different weeks, examining how the feedback addressed knowledge gaps and calibration errors.
After reviewing the feedback, participants completed a 12-item questionnaire on a 5-point Likert scale, evaluating the perceived usefulness of each feedback component: \textit{Feed Up} (Q1--Q4), \textit{Feed Back} (Q5--Q8), and \textit{Feed Forward} (Q9--Q12).

\begin{table}[b!]
\vspace{-8pt}
\caption{\textbf{Feedback quality evaluation results.} 
Participants rated 12 items on a 5-point Likert scale 
(1=not useful, 5=very useful), organized by three feedback 
components: \textit{Feed Up} (Q1--Q4), \textit{Feed Back} 
(Q5--Q8), and \textit{Feed Forward} (Q9--Q12).}
\vspace{-8pt}
\centering
\setlength{\tabcolsep}{5pt}
\resizebox{\textwidth}{!}
{
\begin{tabular}{cl|cc|cc}
\toprule
\midrule
\multirow{2}{*}{} & \multirow{2}{*}{\textbf{Item}} & \multicolumn{2}{c|}{\textbf{Student} ($n$=36)} & \multicolumn{2}{c}{\textbf{Instructor} ($n$=11)} \\
\cmidrule{3-6}
& & $M$ $\uparrow$ & $SD$  & $M$ $\uparrow$ & $SD$  \\
\midrule
\multirow{4}{*}{\rotatebox{90}{\textit{Feed Up}}}
& Q1. Learning position information & \textbf{4.36} & 0.80 & 4.27 & 0.79 \\
& Q2. Learning goals$^\dagger$ & 3.78 & 1.05 & 4.36 & 0.81 \\
& Q3. Position-goal connection$^\dagger$ & 3.94 & 0.98 & \textbf{4.55} & 0.52 \\
& Q4. Priority presentation$^\dagger$ & 3.72 & 1.21 & 4.45 & 0.69 \\
\midrule
\multirow{4}{*}{\rotatebox{90}{\textit{Feed Back}}}
& Q5. Correct knowledge concepts & 4.39 & 0.77 & 4.36 & 0.67 \\
& Q6. Incorrect knowledge concepts & \textbf{4.69} & 0.62 & 4.55 & 0.82 \\
& Q7. Related past errors & 4.53 & 0.65 & \textbf{4.64} & 0.67 \\
& Q8. Four-category classification & 4.06 & 1.01 & 4.36 & 0.67 \\
\midrule
\multirow{4}{*}{\rotatebox{90}{\textit{Forward}}}
& Q9. Priority relearning concepts & 4.44 & 0.84 & \textbf{4.73} & 0.47 \\
& Q10. Review concepts & \textbf{4.53} & 0.77 & \textbf{4.73} & 0.47 \\
& Q11. Challenge concepts$^\dagger$ & 4.11 & 1.01 & \textbf{4.73} & 0.47 \\
& Q12. KM strategy suggestions & 3.86 & 1.17 & 4.45 & 0.52 \\
\midrule
& \textbf{Overall} & 4.20 & -- & 4.52 & -- \\
\midrule
\bottomrule
\multicolumn{6}{l}{\small $^\dagger p < .10$ (Mann-Whitney U test, Student vs. Instructor)}
\end{tabular}}
\label{tab:feedback_eval}
\end{table}

\boldparagraph{Results}
Table~\ref{tab:feedback_eval} presents the feedback quality evaluation results. Overall, both students ($M$=4.20) and instructors ($M$=4.52) rated the feedback positively, indicating general acceptance of the adaptive feedback approach. Among the three components, \textit{Feed Back} received the highest ratings from students. Particularly, the presentation of incorrect knowledge concepts (Q6, $M$=4.69) and related past errors (Q7, $M$=4.53) were highly valued. These findings suggest that learners appreciate concrete information about their knowledge gaps and connections to previously encountered difficulties. For \textit{Feed Forward}, priority relearning concepts (Q9) and review concepts (Q10) were also rated highly by both groups, suggesting that learners appreciate concrete, actionable guidance about what to study next. In contrast, items involving more abstract metacognitive concepts, such as knowledge monitoring strategy suggestions (Q12, $M$=3.86) and the SDT-based four-category classification (Q8, $M$=4.06), received comparatively lower ratings from students, possibly reflecting the unfamiliarity of these concepts despite their overall positive reception.

\boldparagraph{Student-instructor comparison} Instructors consistently rated feedback components higher than students, with marginal significance observed for several \textit{Feed Up} items (Q2--Q4) and one \textit{Feed Forward} item (Q11). This pattern suggests that instructors may recognize more educational value of goal-oriented and forward-looking feedback elements.

\section{Conclusion}
\vspace{-4pt}

We present the Capture-Calibrate-Coach (3C) framework for knowledge monitoring estimation and adaptive feedback. By constructing heterogeneous graphs from learners' open-ended self-reports and formulating latent perceived state inference as link prediction, our approach enables comprehensive assessment of both knowledge states and knowledge monitoring ability. We classify learners into five metacognitive patterns and deliver personalized feedback addressing knowledge gaps and calibration errors. Experiments validate the effectiveness of our HGNN-based inference method, and a user study confirms that the adaptive feedback is perceived as useful by both learners and instructors. A limitation is the evaluation on a single course domain; future work will investigate cross-domain generalization and long-term learning effects.


\begin{credits}
\subsubsection{\ackname} This work was supported by JST CREST Grant Number JPMJCR22D1 and JSPS KAKENHI Grant Numbers JP22H00551, JP25K21360, and JP24K16759.

\end{credits}

\bibliographystyle{splncs04}
\bibliography{reference}

\end{document}